\def\year{2021}\relax
\documentclass[letterpaper]{article} % DO NOT CHANGE THIS
\usepackage{aaai21}  % DO NOT CHANGE THIS
\usepackage{times}  % DO NOT CHANGE THIS
\usepackage{helvet} % DO NOT CHANGE THIS
\usepackage{courier}  % DO NOT CHANGE THIS
\usepackage[hyphens]{url}  % DO NOT CHANGE THIS
\usepackage{graphicx} % DO NOT CHANGE THIS
\usepackage{natbib}  % DO NOT CHANGE THIS AND DO NOT ADD ANY OPTIONS TO IT
\usepackage{caption} % DO NOT CHANGE THIS AND DO NOT ADD ANY OPTIONS TO IT
\usepackage{booktabs} % for professional tables
\usepackage{tabularx}
\usepackage{arydshln}
\usepackage{wrapfig}
\usepackage{amsmath}
\usepackage{amsfonts}
\usepackage{subcaption}
\usepackage{algorithm}
\usepackage{algorithmic}
\usepackage{latexsym}
\usepackage{multirow}
\usepackage{tablefootnote}
\usepackage{longtable}

\begin{document}
% \title{DLGNet: A Transformer-based Model for Dialogue Response Generation}
\title{DLGNet-Task: An End-to-end Neural Network Framework for Modeling Multi-turn Multi-domain Task-Oriented Dialogue}
% \author{Anonymous Authors}
\author{Oluwatobi O. Olabiyi, \textsuperscript{1} 
Prarthana Bhattarai, \textsuperscript{2}
C. Bayan Bruss, \textsuperscript{2} 
Zachary Kulis \textsuperscript{1}\\
\textsuperscript{1} Capital One - Conversation Research, McLean, VA\\
\textsuperscript{2} Capital One - Center for Machine Learning, McLean, VA\\
\{oluwatobi.olabiyi, prarthana.bhattari, bayan.bruss, zachary.kulis\}@capitalone.com
}
% \author{Oluwatobi O. Olabiyi \ and 
% Erik T. Mueller\\
% Capital One Conversation Research, Vienna, VA\\
% \{oluwatobi.olabiyi, erik.mueller\}@capitalone.com
% } %\usepackage{aaai19}  %Required

\iffalse
%Example, Multiple Authors, ->> remove \iffalse,\fi and place them surrounding AAAI title to use it
\title{My Publication Title --- Multiple Authors}
\author {
    % Authors

        First Author Name,\textsuperscript{\rm 1}
        Second Author Name, \textsuperscript{\rm 2}
        Third Author Name \textsuperscript{\rm 1} \\
}
\affiliations {
    % Affiliations
    \textsuperscript{\rm 1} Affiliation 1 \\
    \textsuperscript{\rm 2} Affiliation 2 \\
    firstAuthor@affiliation1.com, secondAuthor@affilation2.com, thirdAuthor@affiliation1.com
}
\fi

\maketitle
\begin{abstract}
Task oriented dialogue (TOD) requires the complex interleaving of a number of individually controllable components with strong guarantees for explainability and verifiability. This has made it difficult to adopt the multi-turn multi-domain dialogue generation capabilities of streamlined end-to-end open-domain dialogue systems. In this paper, we present a new framework, DLGNet-Task, a unified task-oriented dialogue system which employs autoregressive transformer networks such as DLGNet and GPT-2/3 to complete user tasks in multi-turn multi-domain conversations. Our framework enjoys the controllable, verifiable, and explainable outputs of modular approaches, and the low development, deployment and maintenance cost of end-to-end systems. Treating open-domain system components as additional TOD system modules allows DLGNet-Task to learn the joint distribution of the inputs and outputs of all the functional blocks of existing modular approaches such as, natural language understanding (NLU), state tracking, action policy, as well as natural language generation (NLG). Rather than training the modules individually, as is common in real-world systems, we trained them jointly  with appropriate module separations. When evaluated on the MultiWOZ2.1 dataset, DLGNet-Task shows comparable performance to the existing state-of-the-art approaches. Furthermore, using DLGNet-Task in conversational AI systems reduces the level of effort required for developing, deploying, and maintaining intelligent assistants at scale.

% significant improvement over existing approaches, and achieves state-of-the-art performance at both the module and system levels.

\end{abstract}

\begin{figure*}[t]
%\vskip 0.2in
\begin{center}
% \begin{subfigure}{.6\textwidth}
\centerline{\includegraphics[width=1.0\textwidth]{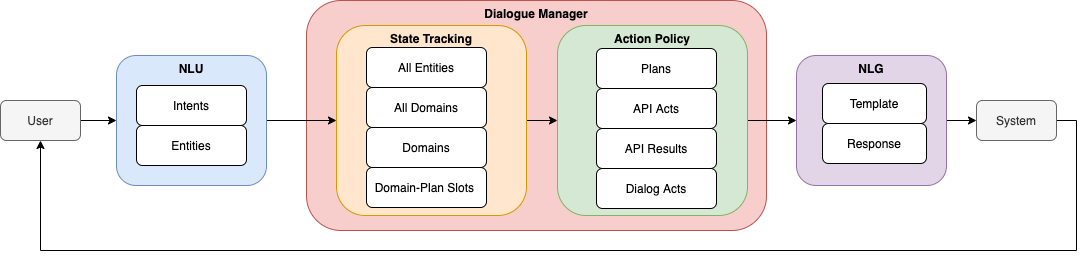}}
\caption{\textbf{DLGNet-Task framework -} The framework captures the end-to-end relationships of all the components of a multi-turn multi-domain task-oriented dialogue system.}
% \source{Figure obtained from \citeauthor{Radford2018} \shortcite{Radford2018}.}
 % hredGAN forces the generator to behave similarly between the training and generative modes.}
\label{dlgnet_module}
\end{center}
\vskip -0.2in
\end{figure*}

\begin{figure*}[ht]
%\vskip 0.2in
\begin{center}
% \begin{subfigure}{.6\textwidth}
\centerline{\includegraphics[width=1.0\textwidth]{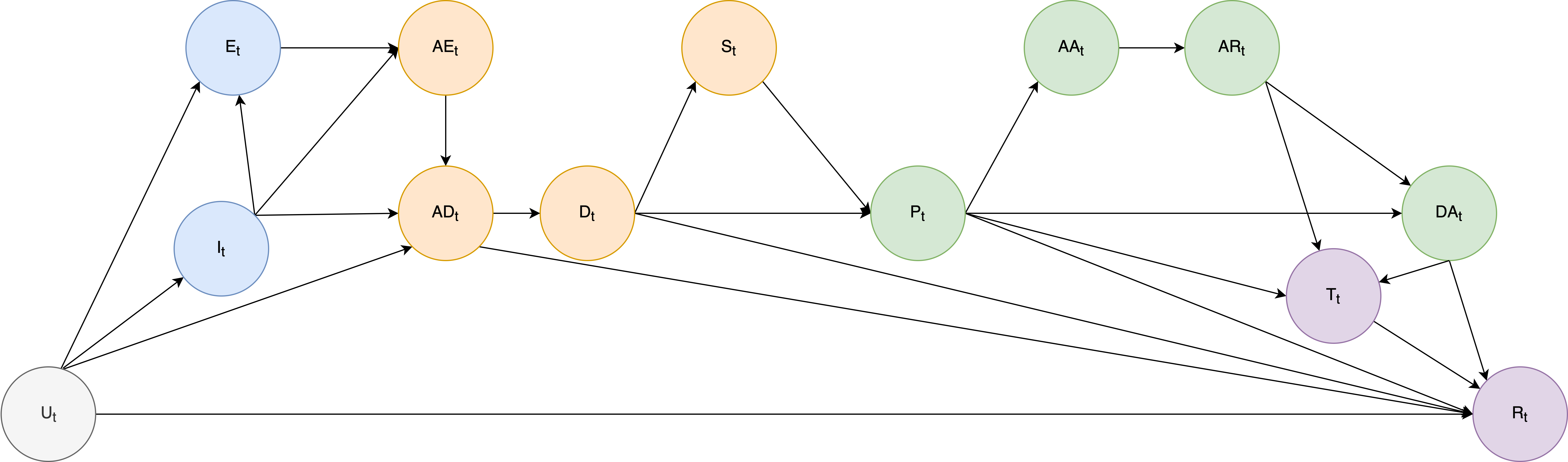}}
\caption{\textbf{The graph of the dialogue flow modeled by DLGNet-Task -} DLGNet-Task models the joint distribution of multiple random variables that exist in both open-domain and task-oriented dialogue.}
% \source{Figure obtained from \citeauthor{Radford2018} \shortcite{Radford2018}.}
 % hredGAN forces the generator to behave similarly between the training and generative modes.}
\label{dlgnet_graph}
\end{center}
\vskip -0.2in
\end{figure*}

\section{Introduction}
\label{introduction}
The desire for human-like interfaces to technical systems, as evidenced by growing use of intelligent assistants, belies the need for conversational AI systems that can accomplish a wide range of tasks, such as booking restaurants, trains, and flights, IT help desk and accessing financial accounts and transaction records. The wide range of tasks have necessitated the need for a flexible and scalable dialogue system that can support a variety of use cases with minimal development and maintenance effort. Existing dialogue systems are broken into two major categories,  open-domain dialogue systems, which focus on non-task related conversations, and task-oriented dialogue systems, which focus on user task completion. A typical open-domain system uses an end-to-end neural architecture often trained with input and output utterances from human-to-human conversations \cite{Sutskever2014,Serban2016,Serban2017,Olabiyi2018,Olabiyi2019,Olabiyi2019c,Zhang2019b}. While open-domain systems are optimized for engaging in human-like conversation, they lack any inherent ability to interface with any other systems on behalf of their conversation partner. Whereas, a typical task-oriented dialogue system seeks to understand human intents and execute them. This is done by adopting a modularized pipeline architecture with three modules that are sequentially connected as shown in Fig. \ref{dlgnet_module}. A natural language understanding (NLU) module that recognizes user intents and extract useful entity information \cite{Kim2017,Lee2019}. The dialogue management (DM) module contains two submodules, the dialogue state tracker (DST) and the dialogue action policy (POL) modules. The DST module tracks the mapping of entities to slots that are relevant or required for completing user tasks \cite{Williams2013}. The POL module decides which actions to execute via the API. Finally, the natural language generation (NLG) module generates the user response based on the user aspects of the system actions \cite{Wen2015}. In some cases, multiple modules are combined together, e.g. systems with a composite NLU and DST module \cite{Ramadan2018,Lee2019b}, and systems with a composite POL and NLG module that maps previous utterances and dialogue states to the system response \cite{Budzianowski2018,Pei2019,Chen2019,Mehri2019,Zhao2019}.

Despite research advances in modular neural approaches, they are hardly used in practice. Industrial dialogue systems, though modularized, still use expensive expert driven rule-based heuristics implemented with several lines of codes and hand-crafted templates, and therefore difficult to scale as the number of use cases grows. More recently, there has been a renewed effort to apply a single end-to-end neural architecture \cite{Budzianowski2019,Ham2020,Peng2020,Hosseini-Asl2020} to model task-oriented dialogue with the use of autoregressive transformer architecture \cite{Radford2019}. This has led to the reformulation of dialogue system design as a text generation or sequence modeling task. While some of these efforts have obtained state-of-the-art performance on publicly available task-oriented dialogue datasets, there is still room for improvement, especially in the areas of generality and practicality. First, their problem formulation fails to reconcile open-domain and task-oriented dialogue in the same model architecture. Also, in many cases, they do not address the complexity of the action policy especially towards the back-end API system. Finally, they don't fully incorporate the control, verification and explanation capabilities that make modularized approaches attractive.

To resolve these shortcomings, we propose DLGNet-Task, an end-to-end neural network that simultaneously handles both open-domain and task-oriented dialogue, in such a way that the model outputs are controllable, verifiable, and explainable at the module level. This system is compatible with both data driven and expert driven rule-based approaches.   That is, our approach is simultaneously modular and end-to-end, and can be a drop-in replacement for traditional modular task-oriented dialogue (TOD) systems. To the best of our knowledge, this is the most expressive approach to date in achieving this objective. In summary, we are able to model the individual behavior of NLU, DM and NLG components with a single neural network model trained end-to-end. Still, the model is flexible enough to allow individual modules to be separately trained and validated in line with the traditional TOD system. 
% Validation at module level can provide information about where additional training is needed. It could also help in balancing the contribution of each module if the model is finetuned with module-level objectives.

% The DLGNet-Task model is based on the autoregressive transformer architecture similar to DLGNet \cite{Olabiyi2019c} and GPT-2/3 \cite{Radford2019,Brown2019} models. To evaluate the performance of DLGNet-Task, we trained the model with just the system-level training objective on a modified MultiWoz2.1 dataset. The dataset modification is done mainly to support DLGNet-Task design framework (example shown in Table \ref{tb:samples_dlgnet}). Based on the widely used TOD metrics, such as inform rate, success rate, and BLEU score \cite{Budzianowski2019}, our experiments show that DLGNet-Task produces a comparable performance to the state-of-the-art approaches on the MultiWoz2.1 dataset. 
% in addition to the controllable, verifiable, and explainable model's intermediate outputs. 

\section{DLGNet-Task System Description}
\label{dlgnet_task}

The goal of DLGNet-Task framework is to support the modeling of both open-domain and task-oriented dialogue. To achieve this, we follow a different modularized pipeline architecture from the ones in recent existing work \cite{Budzianowski2019,Ham2020,Peng2020,Hosseini-Asl2020}. 
DLGNet-Task system consists of (i) the GPT-2 model finetuned using DLGNet objective on a modified MultiWOZ2.1 dataset, (ii) a fine-grained representation of the action policy for calling the backend API and responding to the user, and (iii) intermediate knowledge that allows for controlling, verifying and explaining system behavior.  Fig. \ref{dlgnet_module} shows the functional blocks of the DLGNet-Task framework, which can be applied to any TOD dataset. Fig. \ref{dlgnet_graph} shows the graph of the information flow between the functional blocks in Fig. \ref{dlgnet_module}.  Because it is trained in a single network the DLGNet-Task model learns the joint distribution of the functional blocks based on the information flow graph in Fig. \ref{dlgnet_graph}.  This gives it additional flexibility across tasks and objectives. For example, the direct connection between the utterance node $U_t$ and the response node $R_t$ allows the model to operate in an open-domain context, while other connections allow it to be task oriented. 

\subsection{Framework Components}

DLGNet-Task is broken into three modules as depicted in Fig. \ref{dlgnet_module}, the Natural Language Understanding (NLU) module, the dialogue management module and the Natural Language Generation (NLG) module. Below we describe the components of each of these modules in more detail.

\subsubsection{NLU Module}
The NLU module contains components for identifying intents and entities in the incoming dialogue. It is worth noting that although our NLU model has access to multi-turn context, it only makes predictions relevant to the current turn of the conversation.

\begin{table*}[ht]
\caption{Example dialogue including ground truth responses and DLGNet-Task generated responses}
\label{tb:dlg_net_task_ex1}
\centering
\begin{tiny}
\setlength{\extrarowheight}{3pt}
\begin{tabular}{p{4cm}p{6cm}p{6cm}}
\hline
\textbf{User Prompt} &
  \textbf{Ground Truth Response} &
  \textbf{DLGNet-Task Response} \\ \hline
I am looking for places to go in the centre of town. &
  I have 44 matches for central area, I can narrow it down by type or entrance prices. &
  Okay. My computer shows forty-four attractions in the centre of town. Can you give me some more details so we can narrow the search? \\
How about a park? &
  that should be Cambridge university botanic gardens	located in bateman street. can i give you the phone number? &
  We have cambridge university botanic gardens. Would you like more information about them? \\
Yes, I'll need the number along with the address and postal code. &
  The phone number is 01223336265. &
  The entrance fee is 4 pounds, phone number is 01223336265 and postcode is cb21jf, and the address is bateman street. \\
I also need the address please. &
  The address is Bateman Street in the centre area of town. &
  Postcode is cb21jf, address is bateman street, and you can reach them on 01223336265. \\
Can I get the postcode too please? &
  The postcode is cb21jf. Do you need anything else? &
  I apologize, the postcode is cb21jf, can I assist with anything else? \\
I'm also looking for a place to stay in the same area as Cambridge university botanic gardens &
  Would you prefer a guesthouse or a hotel? &
  We have 5 places to stay all in the city center. Would you like to hear about some of our other options? \\
A guesthouse please and with a star rating of zero. &
  The only place available is the El Shaddai. Would you like for me to book a room? &
  Cityroomz is a moderately priced guesthouse. Would you like me to book a room for you? \\
Actually could you get me the postcode, their price range, and whether they have free parking? &
  The postcode is cb11eg they offer free parking and is cheaply priced. &
  El Shaddai is in the cheap price range, and their postcode is cb11eg. Is there anything else I can assist you with today? \\
Thank you very much. that should be it. &
  I can also recommend restaurants and arrange transportation. Do you need help with these? &
  Thank you for contacting us and have a nice day. \\
No, thanks. I have everything I need. &
  Ok! Have a great day! &
  Great. Glad to help. \\ \hline
\end{tabular}
\end{tiny}
\end{table*}

\begin{table*}[ht]
\caption{Example ground truth component values and DLGNet-Task predictions}
\label{tb:dlg_net_task_ex2}
\centering
\begin{tiny}
\setlength{\extrarowheight}{2pt}
\begin{tabular}{p{54pt}p{201pt}p{201pt}}
\hline
 &
  \textbf{Ground Truth} &
  \textbf{Model Output} \\ \hline
\textbf{User} &
  Yes, Ill need the number along with the address and postal code. &
  Yes, Ill need the number along with the address and postal code. \\
% \textbf{Entities} &
%   type:park, area:centre &
%   \\
\textbf{Domains} &
  attraction &
  attraction \\
\textbf{Slots Search} &
  domain:attraction, area:centre, name:?, type:park &
  domain:attraction, area:centre, name:?, type:park \\
\textbf{API Actions} &
  domain:attraction, search-execute &
  domain:attraction, search-execute \\
\textbf{Results} &
  domain:attraction, {[}'Choice', 1{]}, {[}'Sample', {[}\{'id': '15', 'address': 'bateman street', 'area': 'centre', 'entrance': '4 pounds', 'name': 'cambridge university botanic gardens', 'phone': '01223336265', 'postcode': 'cb21jf', 'pricerange': 'moderate', 'openhours': 'it opens from 10:00 a.m. to 6:00 p.m. from april to september, from 10:00 a.m. to 5:00 p.m. in february march and october, and from 10:00 a.m. to 4:00 p.m. from november to january', 'type': 'park'\}{]}{]} &
  domain:attraction, {[}'Choice', 1{]}, {[}'Sample', {[}\{'id': '15', 'address': 'bateman street', 'area': 'centre', 'entrance': '4 pounds', 'name': 'cambridge university botanic gardens', 'phone': '01223336265', 'postcode': 'cb21jf', 'pricerange': 'moderate', 'openhours': 'it opens from 10:00 a.m. to 6:00 p.m. from april to september, from 10:00 a.m. to 5:00 p.m. in february march and october, and from 10:00 a.m. to 4:00 p.m. from november to january', 'type': 'park'\}{]}{]} \\
\textbf{DLG Actions} &
  domain:attraction, search-noerror-inform:{[}{[}'Phone', '01223336265'{]}{]} &
  domain:attraction, search-noerror-inform:{[}{[}'Addr', 'bateman street'{]}, {[}'Phone', '01223336265'{]}, {[}'Post', 'cb21jf'{]}, {[}'Fee', '4 pounds'{]}{]} \\
\textbf{Delexicalized Output} &
  the phone number is {[}attraction\_phone{]} . &
  the entrance fee is {[}value\_count{]} pounds , phone number is {[}attraction\_phone{]} and postcode is {[}attraction\_postcode{]} , and the address is {[}attraction\_address{]} . \\
\textbf{System Output} &
  The phone number is 01223336265. &
  The entrance fee is 4 pounds, phone number is 01223336265 and postcode is cb21jf, and the address is bateman street. \\ \hline
\end{tabular}
\end{tiny}
\end{table*}

\textbf{\emph{Intent Recognition:}}
%TODO fill out something on intent recognition %
The intent recognition maps the customer utterance to a label. 
% in Table \ref{tb:samples_dlgnet}, 
For example, in Table \ref{tb:samples_dlgnet}, `` I would like to book a reservation at Caffe Uno'' would be mapped to the label \textit{request\_booking}.
One of the limitations of the existing NLU module for natural multi-turn dialogue is the requirement to label every utterance within a conversation even when it is not appropriate. 
For example, mapping ``I am looking for information in Cambridge.'' to a label might be unnecessary. In this case, it is more beneficial to focus on giving an appropriate response rather than the intent recognition.

\textbf{\emph{Entity Recognition:}}
In our framework, we separate named entity recognition (NER) from slot filling in order to be compatible with existing industrial modularized pipeline architecture. Our NER component assigns value to an entity without identifying the domain. Not predicting the domain at this stage makes the NER compatible with both conversational on non-conversational datasets.  

\subsubsection{Dialogue Management Module}
The dialogue management module has two broad functions, state tracking and action policy. Each of these functions are comprised of four components. State tracking maintains \textit{All Entities}, \textit{All Domains}, \textit{Domains}, and \textit{Domain-Plan Slots}, while the Action Policy maintains \textit{Plans}, \textit{API Actions}, \textit{API Results}, and \textit{Dialogue Actions}.

\textbf{\emph{\underline{State Tracking}}}

% TODO give a brief overview of what State Tracking is
The state tracking handles the multi-turn understanding by mapping the context to a well defined representation of the dialogue system ontology e.i., \texttt{DOMAIN, PLAN, SLOT, ENTITY, VALUES, CONSTRAINTS}. The state tracking identifies all the entities and domains in the dialogue history, predicts the active domain, and fill the slots of the active domain using the entity information. This arrangement allows for the verification of slot values with the options and constraints provided in the system ontology.

\textbf{\emph{All Entities:}}
We introduce an \textit{all entities} information node that adds the currently recognized entities (if any) to the previous all entity state i.e., $AE_t = AE_{t-1} + E_t$. With this, we can easily verify or replace the generated \textit{all entities} at any conversation turn.

\textbf{\emph{All Domains:}}
We also introduce an \textit{all domains} information node that maintains a running list of the domains. This node adds the domain $D_t$ in the current turn (if any) to the previously recognized \textit{all domains} state $AD_{t-1}$, i.e., $AD_t = AD_{t-1} + D_t$. With this, we can easily verify or replace the generated \textit{all domains} values  at any conversation turn.

\textbf{\emph{Domains:}}
Similar to \citet{Ham2020}, we predict active domains at the current turn, which is a subset of the all domains node above. The MultiWOZ dataset consists of five domains, hotel, train, restaurant, hotel, and attraction. We support both single and multi-domain interactions over multiple turns of conversation.

\textbf{\emph{Domain-Plan Slots:}}
The slots are divided into three, informable, requestable and book slots. Informable slots represent user constraints and requestable slots hold additional information that the user wants to obtain. Book slots are used to reserve a place recommended by the system or selected by the user \cite{Ham2020}. In order to generalize the slot types to new use cases, we map these categories to their functions (plans). That is, we map informable and book slots to search and booking slots respectively, indicating what the slots are being used for. The requestable slots remain the slots to hold additional information that the user wants to obtain. We predict the plan slots for each predicted domains.

In order to avoid an open ended generation of the dialogue state as in existing work \cite{Budzianowski2019,Ham2020,Peng2020,Hosseini-Asl2020} and improve generalization to new domains, we provide the model with a list of slots for each plan type, and indicate if each slot is filled or not. During inference, with a new domain or plan type, we only need to provide an appropriate slot list, and the model can easily fill it based on the utterance and entity information.

\textbf{\emph{\underline{Action Policy}}}

% TODO write up about Action Policy
The action policy predicts the sub-domain plans relevant to the current turn of the conversation. It also predicts the API actions, and conditioned on the results of the API action, predicts the dialogue act. 

\textbf{\emph{Plans:}}
For each domain, we predict the relevant plans for the current conversation turn. For the MultiWOZ dataset, we have two main plans with slots (search and booking), and others without slots (welcome, greet, bye, and reqmore).

\textbf{\emph{API Actions:}}
For each plan, we determine the appropriate API actions. In the MultiWOZ dataset, we observe two actions, execute (for both search and booking plans), and retrieve (for booking plan). Therefore, our API action uses the format \texttt{[PLAN-ACTION]} for each domain
as shown in the example in Table \ref{tb:samples_dlgnet}.

\textbf{\emph{API Results:}}
For each API action, we call the plan's API with filled plan type slots. For the MultiWOZ dataset, in the case of search, we return the number for matches, as well as top k results. In our ablation studies, we considered $k = \{1, 3, 5 \}$. For booking, we use the results from the dialog-act and/or the requestable slots in the metadata.

\textbf{\emph{Dialogue Actions:}}
For each plan, we select appropriate action(s) among inform, request, recommend, select, book, offerbook, offerbooked. We also identify three status codes, nobook (booking error) and nooffer (search error) and noerror (otherwise). The errors were originally identified as actions in the MultiWOZ dataset, but we assign the corresponding actions as inform in our modification.  Therefore, our dialogue action uses the format \texttt{[PLAN-STATUSCODE-ACTION]} for each domain with appropriate slot information
as shown in the example in Table \ref{tb:samples_dlgnet}.  

Also, we noticed that booking a train requires an offer confirmation before the system can make the booking. This behavior may be required in commercial applications where customer confirmation may be legally required before creating an order or executing a particular plan. To provide a trigger signal to the model, we added a `confirm' slot to the booking slots of train domain 
(see Table \ref{tb:samples_dlgnet}). 
Filling of this slot consequently triggers the `offerbook' action. Since this data modification is an after thought, the entity recognition and slot filling are based on heuristics, which is imperfect and introduces noise into the training data. However, this modification is necessary to fully describe the DLGNet-Task framework.

\subsubsection{NLG Module}

DLGNet-Task handles both template and system response generation. In order to do this, we delexicalized all the values of requestable slots (reference number, name, postcode, phone number, address) as \texttt{[DOMAIN\_SLOTNAME]} (e.g. \texttt{[train\_reference]} for train’s booking reference that appear in the dataset) similar to Ham et al. \shortcite{Ham2020}. Unlike in the existing work, where post processing routine is used to string-replace the delexicalized token later by the real information from the API results, the DLGNet-Task model directly generates the final system response. This makes DLGNet-Task a truly end-to-end neural TOD system.

% \subsection{Discrimination during decoding}
\iftrue
Since DLGNet-Task model is a word-token sequence generation model, the traditional decoding approach is to explore sequence decoding strategies such as greedy decoding, beam-search decoding, top\_k sampling and top\_p sampling strategies \cite{Holtzman2019,Olabiyi2019c}. However, TOD systems contains both natural language and several ontology-driven key-value pairs, such as graph node-value, intent-value, entity-value, slot-entity, domain-value, plan-value, plan-API action, plan-dialogue action pairs. The ontology-driven key-value pairs provide opportunities for discrimination, since some of the key and possible values may be known \textit{apriori} from the system ontology. Note that the ontology itself is not used during training, it is only used here to ground value generation or selection during inference. For example, given the triples $(C, K_i, \{V_i\}_j^J)$ of context $C$, key $K$ and possible values $V$, we can estimate the likelihood of each possible value $V_i^j$, i.e.,
\begin{align}
P_{\theta}\big(V_i^j |K_i, C\big) = DLGNet([C, DL_{key}, K_{i}, DL_{value}, V_i^j])
 \label{eq:dlgnet_score}
\end{align}
where $DL$'s are delimiter tokens.

The likelihood scores can be used to rank possible values during inference, which would also improve generalization to new key-value pairs. Using the likelihood score in Eq. (\ref{eq:dlgnet_score}), a normalized conditional distribution over the value options can be estimated, i.e.,
\begin{align}
P\big(V_i^j | K_i, C\big) = \frac{\exp\big(\frac{1}{T_i}\log P_{\theta}(V_i^j |K_i, C)\big)}{\sum_j^J \exp\big(\frac{1}{T_i}\log P_{\theta}(V_i^j |K_i, C)\big)}
 \label{eq:dlgnet_prob}
\end{align}
where the hyperparamter $T_i\in(0,1]$ is the decoding temperature.

\fi

\subsection{Input Representation}
DLGNet-Task system is implemented with the use of autoregressive transformer networks, such as GPT-2 and DLGNet. To do this, we need to convert the dialogue data to word tokens using the information flow graph. In the case of the MultiWOZ dataset, `metadata' and `dialog-act' correspond to the current dialogue state, and the current system actions. The dataset also contains the user utterance and system response. 
% Table \ref{tb:samples_dlgnet} depicts an example of our processing the MultiWOZ dataset. 
We introduce a delimiter token for each functional block used in each conversation turn. We also introduce two special delimiter tokens, \texttt{$<$turn\_sep$>$} and \texttt{$<$conversation\_sep$>$} for turn and conversation separation respectively based on the DLGNet framework \cite{Olabiyi2019c}. 

% Finally, the models are implemented, trained, and evaluated using Python and the TensorFlow deep learning framework.

%  In order to archive this, we only accumulate the dialogue state at each turn, unlike existing work, where they accumulate the dialogue history. In addition to the joint modeling of open-domain and task-oriented dialogue, our accumulation of dialogue state fits better with commercial dialogue system applications, where expert driven rule-based DM is traditionally employed.

\subsection{Dialogue Information Flow}
In order to obtain the processed data 
as shown in Table \ref{tb:samples_dlgnet}, 
we adopted a dialogue flow process based on the information flow graph in Fig. \ref{dlgnet_graph}. The model inference follows the same process after training. The procedure for processing the training data and model inference using the dialogue flow is described below:
% Once the model is trained the dialogue flows according to the following process:

Given the current turn utterance $U_t$, and the information from the previous dialogue turn $M_{t-1} = \{U_{t-1}, I_{t-1},E_{t-1},\\AE_{t-1},AD_{t-1},D_{t-1},S_{t-1},P_{t-1},AA_{t-1},AR_{t-1},DA_{t-1},\\T_{t-1},R_{t-1}\}$:

\begin{itemize}
\item \textbf{NLU Functional Blocks}
\begin{enumerate}
\item Predict intent $I_t$ if applicable else skip. 
\item Predict entities $E_t$ if applicable else skip.
\end{enumerate} 

\item \textbf{DST Functional Blocks}
\\ Given the NLU predictions:
\begin{enumerate}
\item Predict all entities $AE_t$ if applicable else skip.
\item Predict all domains $AD_t$ if applicable else skip.
\item Predict active domains $D_t$ if applicable else skip.
\item Given the domain, predict plan slots $S_t$ if applicable else skip.
\end{enumerate} 

\item \textbf{POL Functional Blocks}
\\Given the NLU and DST predictions:
\begin{enumerate}
\item Predict plans $P_t$ if applicable else skip. 
\item Given the plan, predict API actions $AA_t$ if applicable else skip.
\item Given the API actions, obtain API results $AR_t$ if applicable else skip.
\item Given the plan, predict dialogue actions $DA_t$ if applicable else skip.
\end{enumerate}

\item \textbf{NLG Functional Blocks}
\\Given NLU, DST and POL predictions:
\begin{enumerate}
\item Predict delexicalized template $T_t$ if applicable else skip. 
\item Predict system response $R_t$.
\end{enumerate}

\end{itemize}

It is worth mentioning that the introduction of NLU functional blocks, and all entities, domain, plan, and template functional blocks helps to control, verify and explain the system response.

\subsection{Training Objective}
For training DLGNet-Task model, we finetune GPT-2 pretrained model using modified MultiWOZ dataset based on the input representation described above. In our experiments, due to the use of turn and conversation separations, we only use the objective of autoregressive language modeling for joint distribution modeling with random informative padding \cite{Radford2019,Olabiyi2019c}, i.e.,

\begin{align}
L_{DLGNet}(x_1,x_2,\cdots,x_n\big) = \sum_{i} \log P_{\theta}\big(x_i |x_1, \cdots, x_{i-1}\big)
 \label{eq:dlgnet}
\end{align}

\subsection{DLGNet-Task Training}
using the Adaptive Moment Estimation (Adam) stochastic gradient descent algorithm with a learning rate of 0.0001 with a maximum sequence length is 1024. Due to GPU memory limitations, we use a batch size of 2 and accumulate  gradients over 5 iterations, making the effective batch size 10. The models are trained until the training perplexity on the dialogue datasets reaches a steady state.

% \begin{table*}[ht]
% \caption{Context-state-to-response evaluation on MultiWOZ}
% \label{tb:eval_context_response}
% \begin{center}
% \begin{small}
% % \begin{sc}
% \vspace{-10pt}
% \setlength\tabcolsep{12.0pt}
% \begin{tabular}{lcccc}
% \toprule
% \textbf{Model}  &   \textbf{Inform} & \textbf{Success} & \textbf{BLEU} & \textbf{Combined} \\
% % \multirow{3}{*}{\textbf{Models}}  &   \multicolumn{2}{c||}{\textbf{Movie}} & \multicolumn{2}{c}{\textbf{Ubuntu}} \\
%  % & \multicolumn{2}{c}{Informativeness} & \multicolumn{2}{c}{Informativeness} \\    
% % & aBoots\_w\_cat & HRED & aBoots\_w\_cat & VHRED & aBoots\_w\_cat & hredGAN\_w & aBoots\_w\_cat & DAIM \\
% \midrule
% DLGNet-Task (Ours)                  & 72.90 &  54.10 & 18.58  & 82.08 \\
% \bottomrule
% \end{tabular}
% \end{small}
% \end{center}
% % \vspace{-20pt}
% \vspace{-10pt}
% \end{table*}

\begin{table*}[ht]
\caption{Context-result-to-response evaluation on MultiWOZ}
\label{tb:eval_context_response}
\begin{center}
\begin{small}
% \begin{sc}
\vspace{-10pt}
\setlength\tabcolsep{12.0pt}
\begin{tabular}{lcccc}
\toprule
\textbf{Model}  &   $\uparrow$ \textbf{Inform} & $\uparrow$ \textbf{Success} & $\uparrow$ \textbf{BLEU} & $\uparrow$ \textbf{Combined} \\
% \multirow{3}{*}{\textbf{Models}}  &   \multicolumn{2}{c||}{\textbf{Movie}} & \multicolumn{2}{c}{\textbf{Ubuntu}} \\
 % & \multicolumn{2}{c}{Informativeness} & \multicolumn{2}{c}{Informativeness} \\    
% & aBoots\_w\_cat & HRED & aBoots\_w\_cat & VHRED & aBoots\_w\_cat & hredGAN\_w & aBoots\_w\_cat & DAIM \\
\midrule
Baseline \citep{Budzianowski2018}   & 71.29 &  60.94 & 18.80  & 84.93 \\
GPT-2 \citep{Budzianowski2019}      & 70.96 &  61.36 & 19.05  & 85.21 \\
Structured Fusion \citep{Mehri2019} & 82.70 &  72.10 & 16.34  & 93.74 \\
SOLOIST \citep{Peng2020}            & 89.60 &  79.30 & 18.03  & 102.49 \\
DSTC8 Track 1 Winner \citep{Ham2020}& 77.00 &  69.70 & 16.11  & 89.46 \\
DLGNet-Task (Ours)                  & 75.15 &  57.31 & 18.34  & 87.52 \\
\bottomrule
\end{tabular}
\end{small}
\end{center}
% \vspace{-20pt}
\vspace{-10pt}
\end{table*}

\begin{table*}[ht]
\caption{End-to-end evaluation on MultiWOZ}
\label{tb:eval_e2e}
\begin{center}
\begin{small}
% \begin{sc}
\vspace{-10pt}
\setlength\tabcolsep{12.0pt}
\begin{tabular}{lcccc}
\toprule
\textbf{Model}  &   $\uparrow$ \textbf{Inform} & $\uparrow$ \textbf{Success} & $\uparrow$ \textbf{BLEU} & $\uparrow$ \textbf{Combined} \\
\midrule
Structured Fusion \citep{Mehri2019} & 73.80 &  58.60 & 16.90  & 83.10 \\
SOLOIST \citep{Peng2020}            & 85.50 &  72.90 & 16.54  & 95.74 \\
SimpleTOD \citep{Hosseini-Asl2020}  & 84.40 &  70.10 & 15.01  & 92.26 \\
DSTC8 Track 1 Winner \citep{Ham2020}& 73.00 &  62.40 & 16.00  & 83.50 \\
DLGNet-Task (Ours)                  & 72.65 &  56.81 & 15.40  & 80.13 \\
\bottomrule
\end{tabular}
\end{small}
\end{center}
% \vspace{-20pt}
\vspace{-10pt}
\end{table*}

\section{Experiments}
\label{experiments}
In this section, we describe how the DLGNet-Task models, trained using the proposed framework performs on the Multiwoz2.1 dataset under different data and model inference conditions.

\subsection{Evaluation Metrics}
Following \citeauthor{Budzianowski2018} \shortcite{Budzianowski2018}, we evaluate the DGLNet-Task models based on the Inform, Success, and BLEU scores. Inform measures if the system provides a correct entity (inform rate). Success measures the exact matching of answering all the requested information (success rate) and if the answered information matches users’ goal . BLEU evaluates how natural the generated utterance is compared with human readers. A combined score (Combined) is also reported using \texttt{Combined = (Inform + Success) x 0.5 + BLEU} as an overall quality measure, as suggested in \citeauthor{Budzianowski2018} \shortcite{Budzianowski2018}.

\subsection{Model Inference Variants}
For the purpose of evaluation, we consider three different model inference variants.

\subsubsection{Context-Result-to-Response}
Under this condition, we provided the ground truth dialogue history, language understanding, belief states, and action policy. The model generates the dialogue actions, delexicalized outputs as well as the final response. This helps to evaluate the combined performance of the dialogue act policy and the system response generation modules. This variant is similar to the Context-to-Response evaluation in the existing work \cite{Peng2020}.

\subsubsection{Context-State-to-Response}
Under this condition, we provided the ground truth dialogue history, language understanding, and the belief states. The model undergoes a 2-stage generation here. First, the model generates the plan and the API actions. The generated API action is combined with the ground truth belief state and converted to backend API calls. If the generated API action is \textit{search-execute}, a database query result is returned. However, if the generated API action is either of \textit{booking-execute} or \textit{booking-retrieve}, we just copy from the ground truth booking results, since we are not able to reproduce the booking experience in real time. If the corresponding ground truth booking result does not exist, the we just use the model generated results (we hope to explore real time booking experience in the future). In the second stage, the model generates the dialogue actions, delexicalized outputs as well as the final response similar to Context-Result-to-Response above.  This evaluation variant helps to evaluate the combined performance of the action policy (API and dialogue) and the system response modules, which was not explored in the existing work.

\subsubsection{End-to-end}
This evaluation variant also involves performing a 2-stage generation similar to the Context-State-to-Response. However, the model input only consists of ground truth dialogue history and the current user utterance. The model first generates the NLU outputs and everything up to the API actions. The generated belief state and the API action is then used to make backend API calls using the procedure described for the Context-State-to-Response variant. In the second stage, the model generates the dialogue actions, delexicalized outputs as well as the final response similar to Context-Result-to-Response above. This evaluation variant helps to evaluate the combined performance of NLU, DST, POL and NLG modules, which is similar to the  end-to-end evaluation in the existing work.

\subsection{Dataset Variants}
We used three different datasets to explore the support for controllability and explainability. The full version, \texttt{FULL} contains all the nodes shown in the dialogue flow graph. The second version, \texttt{MED} excludes the \textit{all\_entities} and \textit{all\_domain} variables. The final version, \texttt{MIN} further excludes the \textit{plans} from the \texttt{MED} version. We also tried excluding the \textit{delex} variable, but the model consistently performed worse, showing that intermediate template generation contributes positively to response generation performance.

\begin{table*}[!htbp]%{0.5\textwidth}
% \caption{Example of training data for DLGNet-Task models based on the MultiWoz dataset}
\label{tb:results}
% \clearpage
% \onecolumn
% \begin{longtable}{cX}
\begin{minipage}[t]{0.5\linewidth}
\caption{Ablation Studies on the Dataset Variations $(k=5)$}
\label{tb:eval_abl_k5}
% \hspace{}
\begin{center}
\begin{small}
% \begin{sc}
\vspace{-0pt}
\setlength\tabcolsep{4.0pt}
\begin{tabular}{lcccc}
\toprule
\textbf{Model}  &   $\uparrow$ \textbf{Inform} & $\uparrow$ \textbf{Success} & $\uparrow$ \textbf{BLEU} & $\uparrow$ \textbf{Combined} \\
\midrule
\multicolumn{2}{l}{\textbf{Context-result-to-response}} \\
\texttt{FULL} & 72.90 &  54.10 & 18.58  & 82.08 \\
\texttt{MED}  & 71.20 &  56.00 & 18.28  & 81.88 \\
\texttt{MIN}  & 65.90 &  51.30 & 18.07  & 76.67 \\

\multicolumn{2}{l}{\textbf{Context-state-to-response}} \\
\texttt{FULL} & 67.40 &  48.50 & 16.34  & 74.30 \\
\texttt{MED}  & 67.10 &  50.10 & 16.42  & 75.02 \\
\texttt{MIN}  & 65.40 &  51.20 & 16.54  & 75.92 \\

\multicolumn{2}{l}{\textbf{End-to-end}} \\
\texttt{FULL} & 63.30 &  37.50 & 13.97  & 64.37 \\
\texttt{MED}  & 62.20 &  41.50 & 14.34  & 66.20 \\
\texttt{MIN}  & 62.50 &  43.00 & 15.29  & 68.04 \\
\bottomrule
\end{tabular}
\end{small}
\end{center}
% \vspace{-20pt}
\vspace{-10pt}
\end{minipage}%
% \hfill
% \hspace{10pt}
\begin{minipage}[t]{0.5\linewidth}
\caption{Ablation Studies on the Search Result Sample Size (Dataset variant \texttt{MIN})}
\label{tb:eval_abl_min}
\begin{center}
\begin{small}
% \begin{sc}
\vspace{-11pt}
\setlength\tabcolsep{4.0pt}
\begin{tabular}{lcccc}
\toprule
\textbf{Model}  &   $\uparrow$ \textbf{Inform} & $\uparrow$ \textbf{Success} & $\uparrow$ \textbf{BLEU} & $\uparrow$ \textbf{Combined} \\
% \multirow{3}{*}{\textbf{Models}}  &   \multicolumn{2}{c||}{\textbf{Movie}} & \multicolumn{2}{c}{\textbf{Ubuntu}} \\
 % & \multicolumn{2}{c}{Informativeness} & \multicolumn{2}{c}{Informativeness} \\    
% & aBoots\_w\_cat & HRED & aBoots\_w\_cat & VHRED & aBoots\_w\_cat & hredGAN\_w & aBoots\_w\_cat & DAIM \\
\midrule
\multicolumn{2}{l}{\textbf{Context-result-to-response}} \\
$k=1$ & 75.15 &  63.23 & 18.34  & 87.53 \\
$k=3$ & 71.20 &  56.00 & 18.29  & 81.89 \\
$k=5$ & 85.50 &  72.90 & 16.54  & 95.74 \\

\multicolumn{2}{l}{\textbf{Context-state-to-response}} \\
$k=1$ & 75.15 &  63.43 & 17.19  & 86.48 \\
$k=3$ & 67.10 &  50.10 & 16.42  & 75.02 \\
$k=5$ & 85.50 &  72.90 & 16.54  & 95.74 \\

\multicolumn{2}{l}{\textbf{End-to-end}} \\
$k=1$ & 72.65 &  56.81 & 15.40  & 80.13 \\
$k=3$ & 67.70 &  52.50 & 15.39  & 75.48 \\
$k=5$ & 62.50 &  43.00 & 15.29  & 68.04 \\
\bottomrule
\end{tabular}
\end{small}
\end{center}
% \vspace{-20pt}
\vspace{-10pt}
\end{minipage}
\end{table*}

\section{Results and Discussion}
\label{res_dis}
This section describes the experimental results and discusses some of the unique patterns observed when applying the DLGNet-Task framework and model to the MultiWoz dataset.

\subsection{Quantitative Evaluation}
\label{eval_quant}
The performance of DLGNet-Task is reported in Tables \ref{tb:eval_context_response} and \ref{tb:eval_e2e} for context-result-to-response and end-to-end evaluations respectively. DLGNet-Task models produce performance comparable with relevant existing work, although the models do not outperform state-of-art models based on the quantitative measure. The performance gap can mostly be attributed to the noise introduced to the dataset during our processing. The additional control and verification signals introduced to training data might have contributed to the model's reduced performance since they are based on heuristics which might not always be accurate. Also, there are lots of mismatches between the search result and the actual ground truth dialogue actions. These mismatches requires non-trivial effort to fix as also reported by \citet{Hosseini-Asl2020}. To obtain the best result, \citet{Hosseini-Asl2020} did not include the search results in the downstream generation, which is not feasible for practical applications. Also, to obtain the state-of-the-art SOLOIST models, the authors excluded the entire action policy from their framework. However, in practical applications, at least the API action policy is required to know when to execute search, execute booking and retrieve booked information, which is not considered in any of the existing work. DLGNet-Task, on the other hand, does not sacrifice practical usefulness for the purpose of performance on a noisy dataset.

When compared to other practical approaches, such as DSTC8 Track 1 Winner \cite{Ham2020}, DLGNet-Task performs favorably, in addition to the capturing of the API action policy.

\subsection{Qualitative Evaluation}
\label{eval_quanl}
A random conversation sample with the end-to-end model (with $k=1$ and dataset variant \texttt{MIN}) and ground truth outputs are shown in Tables \ref{tb:dlg_net_task_ex1} and \ref{tb:dlg_net_task_ex2}. The tables show DLGNet-Task model being able to follow the multi-turn conversation, correctly executing the dialogue state tracking, and the action policy, and language generation function at each conversation turn. One striking observation is the high level of coherence and relevance in the generated responses from the DLGNet-Task model. The model is also able to capture both short- and long-term temporal dependencies over multiple turns of conversation.

\subsection{Ablation Studies}
\label{ablation}
In our ablation studies we explored the impact of varying the training signals in the dataset, and the size of the search results from the backend API, on the model performance.

\subsubsection{Effect of training signals with dataset variation}
Table \ref{tb:eval_abl_k5} shows the performance of DLGNet-Task model on the dataset with varying amount of training signals with a fixed search result sample size $k=5$. It could be observed that for all dataset varations, the models performs worse with decreasing ground truth information from context-result-to-response, context-state-to-response, to end-to-end as expected. We also observe that the more training signal we have, the better the model performance with more ground truth information during inference (e.g., context-result-to-response). However, with less ground truth information (e.g., end-to-end),the dataset with the least training signal (\texttt{MIN}) gives the best results. We attribute this surprising behavior to the noise in the original dataset and our data processing. Therefore, another way to look at this is that the model gives the best end-to-end performance with the least noisy data. 
% Since the end-to-end performance is of greater interest, we hope to explore the use of classification of dataset variation as an additional reinforcement signal during training to further improve the model performance in the future. 

\subsubsection{Effect of Search Result Sample Size}
We also compare the model performance with different sizes of the search result sample. The results with best performing the \texttt{MIN} dataset variation are reported in Table \ref{tb:eval_abl_min}. The results show that the model performs worse with increasing result sample size. Although this is also surprising, the behavior could be attributed to the disparity in the informed entities in the ground truth dialogue action policy and the ones in the search result. Even when the informed entities are present in the search result, they are in random positions, which makes it difficult to obtain any useful training signal for the model. This observation shows the need for a consistent training dataset in order to evaluate both the system architecture and the proposed neural network models for end-to-end TOD systems.

\section{Conclusion}
\label{concl}
In this paper, we have proposed DLGNet-Task, an end-to-end neural network framework for modeling multi-turn multi-domain task-oriented dialogue. The DLGNet-Task model learns the joint distribution of the nodes (variables) of a dialogue flow graph capable of representing both task-oriented and open-domain dialogue systems. For TOD specific applications, DLGNet-Task is also capable of learning the action policy towards the back-end API. Experimental results show that DLGNet-Task gives comparable performance with existing approaches with practical focus. The results also showed that performance of DLGNet is hampered by the errors in the original MultiWoz dataset as well as the noise introduced during the data processing. 
% While DLGNet-task framework is capable to learning a controllable, verifiable and explainable end-to-end model. 
This also shows need for consistent TOD datasets with properly defined dialogue flow graph. We hope to explore this direction as part of our future work both in terms of dataset generation and data processing pipeline. We also hope to improve DLGNet-Task model performance with adversarial and reinforcement learning. 

% \newpage
% \clearpage
% \break
\bibliography{aaai_2021_dlgnet_task}
% \bibliographystyle{aaai21}

% \iffalse
\iftrue
\setcounter{table}{0}
\renewcommand{\thetable}{A\arabic{table}}

\newpage
\appendix
% \section{Appendices}
% \label{sec:appendix}

% \begin{table*}[!htbp]%{0.5\textwidth}
\begin{table*}[t]%{0.5\textwidth}
\caption{Example of training data for DLGNet-Task models based on the MultiWoz dataset}
\label{tb:samples_dlgnet}
% \clearpage
% \onecolumn
% \begin{longtable}{cX}
\begin{minipage}[t]{0.5\linewidth}
% \caption{Example of training data for DLGNet-Task models based on the MultiWoz dataset}
% \label{tb:samples_gpt}
%\vskip 0.15in
\vspace{-10pt}
\begin{tiny}
\begin{center}
% \begin{sc}
\setlength\tabcolsep{2.0pt}
% \begin{tabular}{lll}
\begin{tabularx}{0.95\linewidth}{ lX }
\toprule
Prompt & Outputs \\
\midrule
\multicolumn{2}{l}{$<$conversation\_sep$>$} \\
usr: & I am looking for information in Cambridge.\\
all\_domains: & general \\
domains: & general\\
plans: & domain:general, reqmore \\
dlg\_acts: & domain:general, reqmore-noerror-inform \\
delex: & sure , what do you need to know ? \\
sys: & Sure, what do you need to know? \\
\multicolumn{2}{l}{$<$turn\_sep$>$} \\
\midrule
usr: & I would like to book a reservation at Caffe Uno.\\
intents: & request\_booking \\
entities: & name:caffe uno \\
all\_entities: & name:caffe uno \\
all\_domains: & restaurant, general \\
domains: & restaurant \\
slots\_search: & domain:restaurant, food:?, pricerange:?, area:?, name:caffe uno \\
slots\_booking: & domain:restaurant, time:?, day:?, people:? \\
plans: & domain:restaurant, booking \\
dlg\_acts: & domain:restaurant, booking-noerror-request:[['Day', '?'], ['Time', '?'], ['People', '?']] \\
delex: & what day and time do you need the reservation for ? and how many people in your group ? \\
sys: & What day and time do you need the reservation for? And how many people in your group? \\
\multicolumn{2}{l}{$<$turn\_sep$>$} \\
\midrule
usr: & I need the reservation for 1 person for Friday at 10:15. \\
entities: & people:1, day:friday, time:10:15 \\
all\_entities: & people:1, day:friday, time:10:15, name:caffe uno \\
all\_domains: & restaurant, general \\
domains: & restaurant \\
slots\_search: & domain:restaurant, food:?, pricerange:?, area:?, name:caffe uno \\
slots\_booking: & domain:restaurant, time:10:15, day:friday, people:1 \\
plans: & domain:restaurant, booking \\
api\_acts: & domain:restaurant, booking-execute \\
results: & domain:restaurant, booked:[{'name': 'caffe uno', 'reference': '3UH2KQDP'}] \\
dlg\_acts: & domain:restaurant, booking-noerror-book:[['Ref', '3UH2KQDP']] \\
delex: & your booking was successful and your table has been reserved ! your reference number is [restaurant\_reference] . can i do anything else for you today ? \\
sys: & Your booking was successful and your table has been reserved! Your reference number is 3UH2KQDP. Can I do anything else for you today? \\
\multicolumn{2}{l}{$<$turn\_sep$>$} \\
\midrule
usr: & Yes, I need a train on Friday also, going into Cambridge. \\
intents: & request\_train \\
entities: & destination:cambridge, day:friday \\
all\_entities: & destination:cambridge, day:friday, people:1, time:10:15, name:caffe uno \\
all\_domains: & train, restaurant, general \\
domains: & train \\
slots\_search: & domain:train, destination:cambridge, day:friday, departure:?, arriveBy:?, leaveAt:? \\
slots\_booking: & domain:train, confirm:?, people:? \\
plans: & domain:train, search \\
dlg\_acts: & domain:train, search-noerror-request:[['Arrive', '?'], ['Depart', '?']] \\
delex: & ok , where from and what time do you want to arrive by ? \\
sys: & Ok, where from and what time do you want to arrive by? \\
\multicolumn{2}{l}{$<$turn\_sep$>$} \\
\midrule
usr: & I am leaving peterborough and want to get to cambridge by 08:15 \\
intents: & request\_booking \\
entities: & arriveBy:08:15, departure:peterborough \\
all\_entities: & arriveBy:08:15, departure:peterborough, destination:cambridge, day:friday, people:1, time:10:15, name:caffe uno \\
all\_domains: & train, restaurant, general \\
domains: & train \\
slots\_search: & domain:train, destination:cambridge, day:friday, departure:peterborough, arriveBy:08:15, leaveAt:? \\
slots\_booking: & domain:train, confirm:?, people:? \\
plans: & domain:train, search, booking \\
api\_acts: & domain:train, search-execute \\
results: & domain:train, ['Choice', 5], ['Sample', [{'trainID': 'TR0035', 'arriveBy': '08:09', 'day': 'friday', 'departure': 'peterborough', 'destination': 'cambridge', 'duration': '50 minutes', 'leaveAt': '07:19', 'price': '16.50 pounds'}, {'trainID': 'TR8286', 'arriveBy': '07:38', 'day': 'friday', 'departure': 'peterborough', 'destination': 'cambridge', 'duration': '50 minutes', 'leaveAt': '06:48', 'price': '16.50 pounds'}, {'trainID': 'TR3144', 'arriveBy': '07:09', 'day': 'friday', 'departure': 'peterborough', 'destination': 'cambridge', 'duration': '50 minutes', 'leaveAt': '06:19', 'price': '16.50 pounds'}, {'trainID': 'TR1799', 'arriveBy': '06:38', 'day': 'friday', 'departure': 'peterborough', 'destination': 'cambridge', 'duration': '50 minutes', 'leaveAt': '05:48', 'price': '16.50 pounds'}, {'trainID': 'TR1662', 'arriveBy': '06:09', 'day': 'friday', 'departure': 'peterborough', 'destination': 'cambridge', 'duration': '50 minutes', 'leaveAt': '05:19', 'price': '16.50 pounds'}]] \\
dlg\_acts: & domain:train, booking-noerror-offerbook:[['Arrive', '08:09'], ['Id', 'tr0035']], booking-noerror-request:[['People', '?']] \\
delex: & i can put you on the [train\_id] it will get you there by [value\_time] . how many tickets do you need ? \\
sys: & I can put you on the tr0035 it will get you there by 08:09. How many tickets do you need? \\
\multicolumn{2}{l}{$<$turn\_sep$>$} \\
\bottomrule
%\end{tabular}
\end{tabularx}
%\end{sc}
\end{center}
\end{tiny}
%\vskip -0.1in
\end{minipage}%
% \hfill
% \hspace{10pt}
\begin{minipage}[t]{0.5\linewidth}
% \caption{Example of training data for DLGNet-Task models based on the MultiWoz dataset}
% \label{tb:samples_gpt}
\vspace{-10pt}
\begin{tiny}
\begin{center}
% \begin{sc}
\setlength\tabcolsep{2.0pt}
% \begin{tabular}{lll}
\begin{tabularx}{0.95\linewidth}{ lX }
\toprule
Prompt & Outputs \\
\midrule
usr: & Just one.  Can I get the reference number for that? \\
entities: & confirm:yes, people:1 \\ % implicit confirm case
all\_entities: & confirm:yes, people:1, arriveBy:08:15, departure:peterborough, destination:cambridge, day:friday, time:10:15, name:caffe uno \\
all\_domains: & train, restaurant, general \\
domains: & train \\
slots\_search: & domain:train, destination:cambridge, day:friday, departure:peterborough, arriveBy:08:15, leaveAt:? \\
slots\_book: & domain:train, confirm:yes, people:1 \\
plans: & domain:train, booking \\
api\_acts: & domain:train, booking-execute \\
results: & domain:train, booked:[['Ref', 'E75VHN9I '], ['Ticket', '16.5 pounds']] \\
dlg\_acts: & domain:train, booking-noerror-inform:[['Ref', 'E75VHN9I '], ['Ticket', '16.5 pounds']] \\
delex: & your reference number is [train\_reference] and [value\_price] pounds will be payable at the station . anything else i can do for you ? \\
sys: & Your reference number is E75VHN9I and 16.5 pounds will be payable at the station. Anything else I can do for you? \\
\multicolumn{2}{l}{$<$turn\_sep$>$} \\
\midrule
usr: & Yes, could you please email or fax me the fare amount, as well as the reference number? \\
all\_domains: & train, restaurant, general \\
domains: & train \\
all\_entities: & confirm:yes, people:1, arriveBy:08:15, departure:peterborough, destination:cambridge, day:friday, time:10:15, name:caffe uno \\
slots\_search: & domain:train, destination:cambridge, day:friday, departure:peterborough, arriveBy:08:15, leaveAt:? \\
slots\_book: & domain:train, confirm:yes, people:1 \\
slots\_requestable: & domain:train, booked:[{'trainID': 'TR0035', 'reference': 'E75VHN9I'}] \\
plans: & domain:train, booking \\
api\_acts: & domain:train, booking-retrieve \\
results: & domain:train, booked:[['Ticket', '16.50 pounds'], ['Ref', 'E75VHN9I']] \\
dlg\_acts: & domain:train, booking-noerror-inform:[['Ticket', '16.50 pounds'], ['Ref', 'E75VHN9I']] \\
delex: & it will cost you [value\_price] pounds and the reference number is [train\_reference] . \\
sys: & it will cost you 16.50 pounds and the Reference number is : E75VHN9I. \\
\multicolumn{2}{l}{$<$turn\_sep$>$} \\
\midrule
usr: & Thank you. That is all that I need. \\
all\_entities: & confirm:yes, people:1, arriveBy:08:15, departure:peterborough, destination:cambridge, day:friday, time:10:15, name:caffe un \\
all\_domains: & general, train, restaurant \\ 
domains: & general \\
plans: & domain:general, welcome \\
dlg\_acts: & domain:general, welcome-noerror-inform \\
delex: & welcome , it was a pleasure serving you . \\
sys: & Welcome, it was a pleasure serving you. \\
\multicolumn{2}{l}{$<$conversation\_sep$>$} \\

\bottomrule
\end{tabularx}
% \vspace{-20pt}
% \end{sc}
\end{center}
\end{tiny}
%\vskip -0.1in
\end{minipage}
\end{table*}
% \end{longtable}
% \clearpage
% \twocolumn

\fi

\end{document}